\documentclass[pdflatex,sn-mathphys-num]{sn-jnl}

\usepackage{graphicx}%
\usepackage{multirow}%
\usepackage{amsmath,amssymb,amsfonts}%
\usepackage{amsthm}%
\usepackage{mathrsfs}%
\usepackage[title]{appendix}%
\usepackage{xcolor}%
\usepackage{textcomp}%
\usepackage{manyfoot}%
\usepackage{booktabs}%
\usepackage{algorithm}%
\usepackage{algorithmicx}%
\usepackage{algpseudocode}%
\usepackage{listings}%

\begin{document}

\title[Article Title]{Active Learning with Fully Bayesian Neural Networks for Discontinuous and Nonstationary Data}

\author[1]{\fnm{Maxim} \sur{Ziatdinov}}

\affil[1]{\orgdiv{Physical Sciences Division}, \orgname{Pacific Northwest National Laboratory}, \orgaddress{\city{Richland}, \state{Washington}, \country{USA}, \postcode{99354}}}

\abstract{
Active learning optimizes the exploration of large parameter spaces by strategically selecting which experiments or simulations to conduct, thus reducing resource consumption and potentially accelerating scientific discovery. A key component of this approach is a probabilistic surrogate model, typically a Gaussian Process (GP), which approximates an unknown functional relationship between control parameters and a target property. However, conventional GPs often struggle when applied to systems with discontinuities and non-stationarities, prompting the exploration of alternative models. This limitation becomes particularly relevant in physical science problems, which are often characterized by abrupt transitions between different system states and rapid changes in physical property behavior. Fully Bayesian Neural Networks (FBNNs) serve as a promising substitute, treating all neural network weights probabilistically and leveraging advanced Markov Chain Monte Carlo techniques for direct sampling from the posterior distribution. This approach enables FBNNs to provide reliable predictive distributions, crucial for making informed decisions under uncertainty in the active learning setting. Although traditionally considered too computationally expensive for 'big data' applications, many physical sciences problems involve small amounts of data in relatively low-dimensional parameter spaces. Here, we assess the suitability and performance of FBNNs with the No-U-Turn Sampler for active learning tasks in the 'small data' regime, highlighting their potential to enhance predictive accuracy and reliability on test functions relevant to problems in physical sciences.}


\maketitle

\section{Introduction}\label{sec1}

Active learning \cite{cohn1996active, settles2009active} is a machine learning technique for efficiently exploring large parameter spaces. By strategically choosing which experiments or simulations to perform next, active learning algorithms can significantly reduce the number of observations needed to navigate these spaces effectively. This not only has the potential to speed up the discovery process but also to minimize resource use, making it a valuable approach in fields where experimental costs are high or when the experimental setup is particularly complex.

The underlying principle is straightforward. We identify a space of experimental or simulation parameters that we can control. For each combination of these parameters, we can measure an associated physical property. We assume the existence of a function that links these parameters to the target property, but its exact form is unknown — if it were known, machine learning would be unnecessary. Instead, we approximate it with a probabilistic surrogate model. Given that it's usually impractical to explore every possible combination of parameters due to resource constraints, and random exploration isn't efficient, we rely on the probabilistic predictions from this surrogate model to intelligently guide our selection of next measurements.

Gaussian Processes (GP) are often the default choice for the surrogate model. They are prized for their capacity to provide uncertainty quantification and smooth interpolations, which are essential in active learning contexts where the goal is to reconstruct unknown ("black box") functions across a parameter space with limited data.  Standard GP assumes the target function is smooth and continuous, and their effectiveness relies on this assumption. As a result, they struggle with discontinuous or non-stationary functions, where abrupt changes or variable statistical properties across the input space challenge the model's basic premises.

Several techniques have been developed to mitigate this limitation. For instance, incorporating a change point model \cite{saatcci2010changepoint, pmlr-v162-caldarelli22a-changepoint} can effectively address the shortcomings of GPs in modeling functions with abrupt transitions between different states. However, this approach depends on sophisticated heuristics and assumptions about the specific type of processes involved, making it challenging to apply universally to arbitrary physical systems, especially in active learning scenarios, where the initial dataset is minimal and the underlying data behavior is largely unknown.

Introducing a prior mean function, whether probabilistic \cite{ziatdinov_hypothesis} or deterministic \cite{noack2021gaussian}, into a GP is another approach to address the above limitations. This prior mean function acts as a baseline expectation of the system's behavior, guiding the GP towards more plausible values based on prior knowledge or theoretical insights about the function being modeled. For instance, if some basic physical laws or trends are known, these can be encoded within the prior mean to inform the GP's predictions, potentially improving both the efficiency and accuracy of the learning process. However, challenges arise when the physical model of the system is unknown or when the relationships within the data cannot be easily expressed in a learnable (i.e. differentiable) form suitable for integration into a GP. In such cases, specifying an appropriate prior mean function becomes non-trivial. If the system's behavior is highly complex or the underlying physics is not well-understood, the use of an incorrect prior mean can lead to biased or misleading predictions.

\begin{figure}[h] 
  \centering  
  \includegraphics[width=1.0\textwidth]{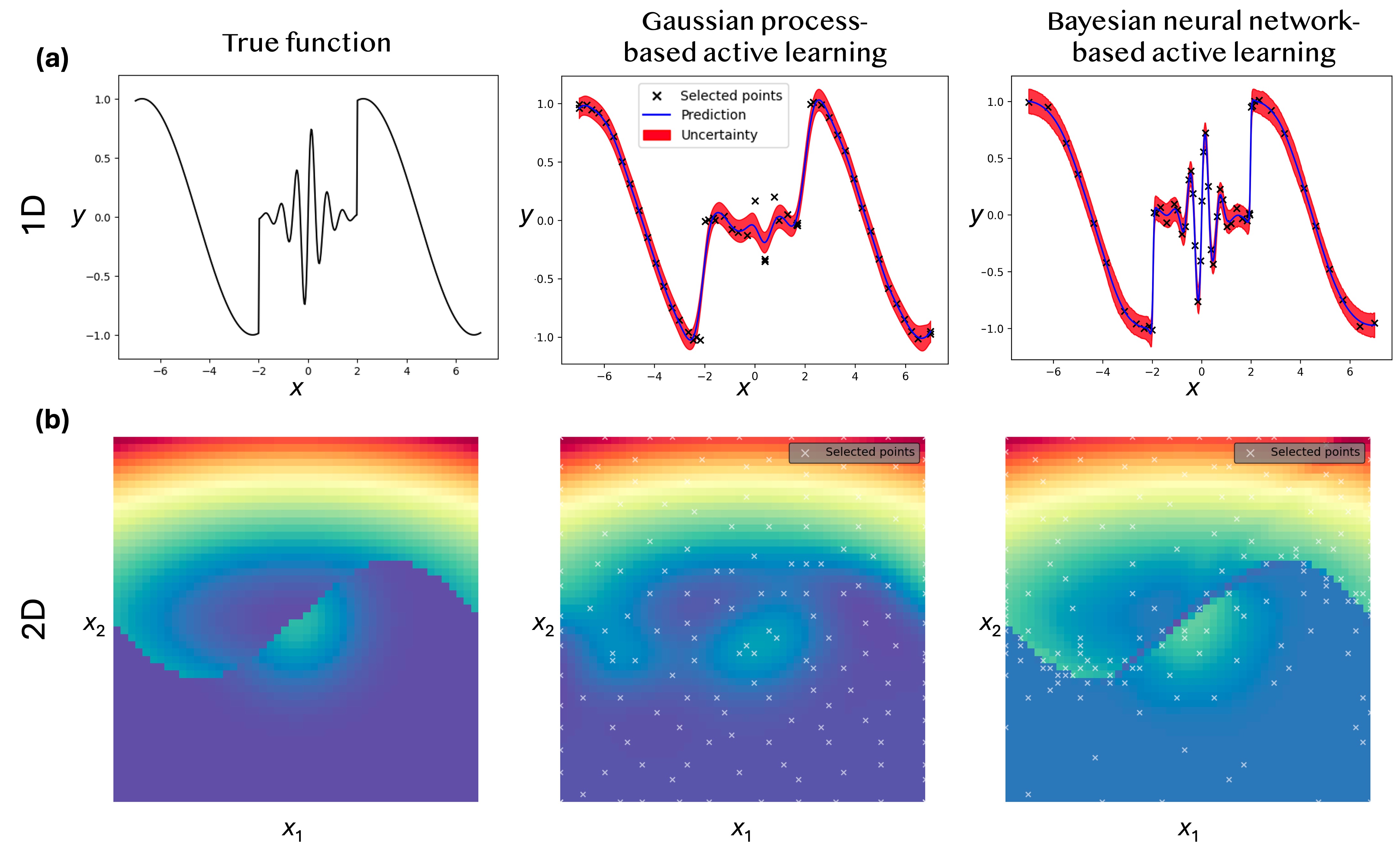}
  \caption{Fully Bayesian neural networks can outperform their Gaussian process counterparts on active learning tasks, particularly for systems exhibiting rapid changes in behavior and/or discontinuous transitions between different states. These types of behavior are often encountered in various physical science domains.}
  \label{fig:Figure1}  
\end{figure}

In the absence of prior knowledge, deep kernel learning (DKL), which combines a neural network with a GP, emerged as a promising strategy for overcoming some of the limitations of traditional GPs \cite{wilson2015deep}. In DKL, the neural network transforms the input data into a new feature space that is more amenable to GP. Thus far, most DKL work has focused on problems in high-dimensional input spaces using either Maximum a Posteriori or Maximum Likelihood Estimation approximations \cite{liu2022experimental, coarsegrained_dkl}. However, Ober et al. highlighted that standard DKL might be prone to overfitting issues \cite{ober2021dklpromises} and proposed that a fully Bayesian approach — encompassing probabilistic treatments of both neural network weights and GP hyperparameters — could mitigate these issues. However, considering that DKL was initially conceptualized as an approximation of a Bayesian deep learning, it raises the question of whether directly employing a fully Bayesian neural network might be more effective.

A Fully Bayesian Neural Network (FBNN) employs a probabilistic treatment of all neural network weights, treating them as random variables with specified prior distributions \cite{neal1996bnn, andrieu1999robust}. It utilizes advanced Markov Chain Monte Carlo techniques to sample directly from the posterior distribution of its weights, allowing the FBNN to account for all plausible weight configurations. This approach enables the network to make probabilistic predictions, not just single-point estimates but entire distributions of possible outcomes, quantifying the inherent uncertainty. Despite their theoretical advantages in providing robust uncertainty quantification, FBNNs have not been widely adopted due to their significant computational demands \cite{izmailov2021bayesian}. Generally, research in Bayesian deep learning has been geared towards finding suitable approximations to FBNNs for handling large, high-dimensional data problems. Yet, in the physical sciences, especially in the experimental research, where parameter spaces are often relatively lower-dimensional, a fully Bayesian approach could be feasible. Employing FBNNs for active learning in these settings could provide reliable uncertainty estimates and robust predictive modeling, essential for scientific applications. Despite these potential advantages, there has been limited effort to systematically evaluate the performance of FBNNs against benchmarks relevant to problems in the physical sciences.

Our objective is to assess and compare the performance of FBNNs across various active learning tasks relevant to autonomous scientific discovery. The central question we explore is: "Should you, as a domain scientist, consider a FBNN for your experimental or modeling needs?" To investigate this, we have developed test functions that are pertinent to common challenges in the physical sciences, such as phase transitions, and used them to rigorously assess the comparative strengths and weaknesses of FBNNs against traditional GPs. Our results show that FBNNs either outperform or at the very least show similar performance to traditional GPs, making them a compelling option for practitioners exploring active learning methods in experimental or modeling applications.

\section{Related work}
To date, active learning with BNNs typically involved various approaches to approximate inference of BNN weights, aimed at balancing computational efficiency with the accuracy of uncertainty estimates. These approaches include Monte Carlo Dropout, Deep Ensembles, and Variational Inference.

Monte Carlo (MC) dropout approach uses dropout at test time to approximate posterior sampling, providing a practical yet rough approximation of Bayesian inference and enabling crude uncertainty estimation in neural network predictions \cite{gal2016dropout}. Y. Gal \textit{et al.} \cite{gal2017deep} first showed its application for active learning on image classification tasks.  J.O. Woo further employed the MC dropout technique with a balanced entropy acquisition method for improved uncertainty estimates on active learning for categorical prediction tasks \cite{woo2023active}. In related work, Fronzi et al. demonstrated the use of MC dropout-based active learning and density functional theory to efficiently predict bandgaps in novel Van der Waals heterostructures \cite{BNN_vdW}. However, careful calibration of dropout rates and neural network architecture design are critical to achieve reliable uncertainty estimates with this method \cite{verdoja2020dropoutnotes}.

Deep ensembles involve training multiple models independently and using the variance in their predictions to estimate uncertainty \cite{lakshminarayanan2017ensembles, deep_ensembles}. While not inherently Bayesian, this method provides a practical way to gauge model confidence and can capture model disagreement effectively. Recent studies have shown that deep ensembles outperform MC Dropout in terms of providing more reliable and well-calibrated uncertainty estimates, thereby improving the active learning process for classification tasks \cite{deepensembles, pop2018deepensembles}. Further extending this approach, the work by Jesús Carrete \textit{et al.} \cite{deep_ensembles_MD} illustrates how deep ensembles can be adapted to improve uncertainty estimation in computational chemistry, enhancing active learning workflows for molecular dynamics simulations.

Variational Inference (VI) aims at finding the best parameters for the assumed distribution that approximates the true posterior \cite{vi}. This method is computationally efficient but may not capture the full complexity of the posterior distribution. A notable application of VI in neural networks is Bayes by Backprop (BBB), which maintains a distribution (usually Gaussian) over each weight and using backpropagation to update the parameters of this distribution, allowing in principle learning the uncertainty in the weights.  Hafner \textit{et al.} \cite{hafner2019bbbnoise} showed that integrating BBB with noise contrastive priors enhances uncertainty estimation for active learning tasks by addressing out-of-distribution data, improving the model's ability to select informative samples and preventing overfitting.

The advanced MCMC methods such as Hamiltonian Monte Carlo (HMC) are considered fully Bayesian because they attempt to sample directly from the posterior distribution. These methods are computationally intensive but provide the most accurate representation of uncertainty by thoroughly integrating over all model uncertainties. Li et al. has recently investigated FBNNs utilizing HMC sampling for Bayesian optimization demonstrating promising results across various benchmark datasets \cite{li2023bnnbo}. Our focus, however, is on evaluating FBNNs for active learning, rather than Bayesian optimization, as exploring and understanding the overall parameter space is a more prevalent objective in scientific research aimed at discerning how physical properties or functionalities behave under varying conditions.

\section{Methods}

\subsection{Fully Bayesian Neural Networks}

\begin{figure}[h] 
  \centering  
  \includegraphics[width=1.0\textwidth]{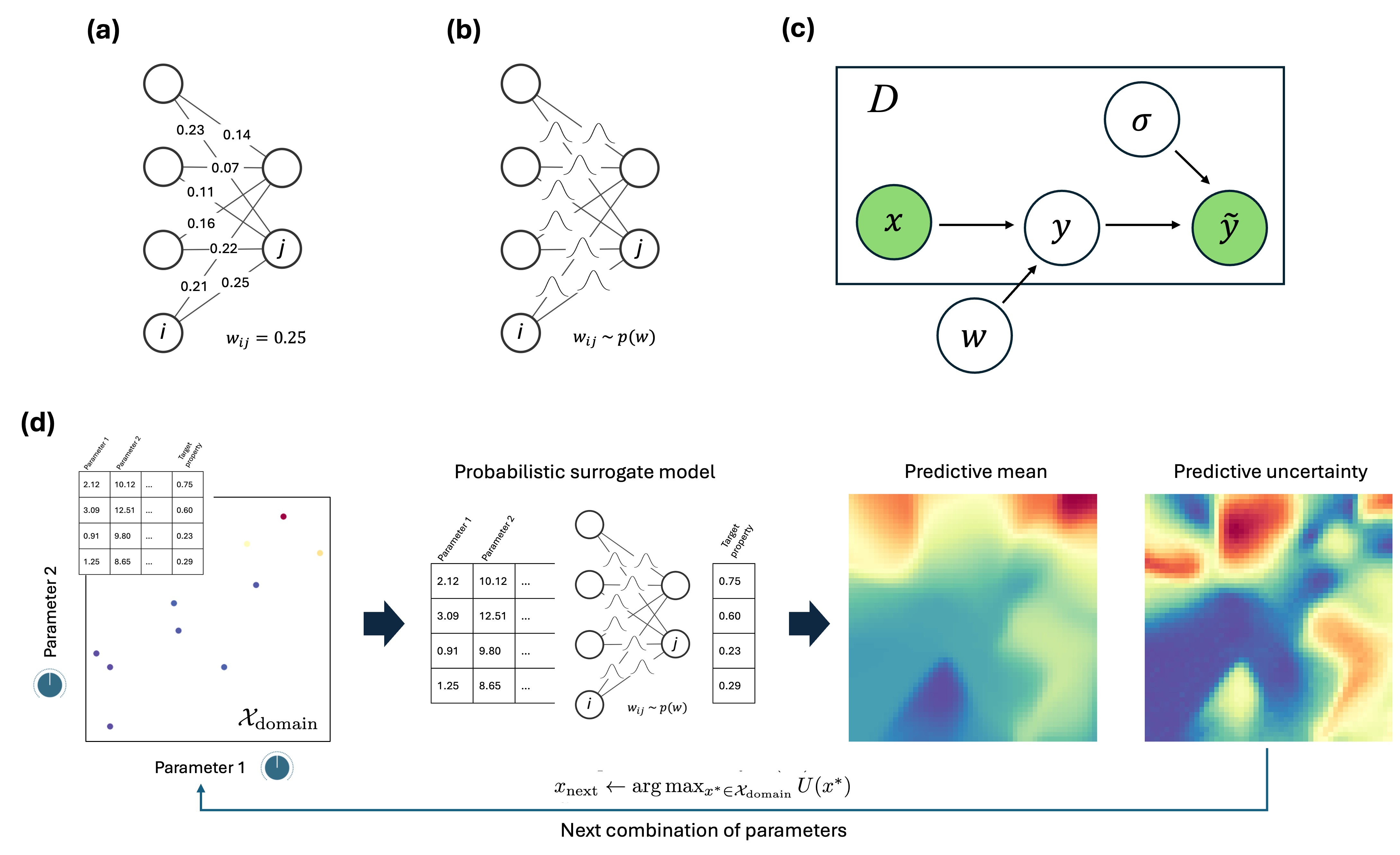}
  \caption{Fully Bayesian neural network (FBNN). The deterministic weights of traditional neural networks (a) are replaced with probabilistic distributions in FBNNs (b). (c) FBNN as probabilistic graphical model: colored circles depict observed variables while unobserved variables are in empty circles. (d) Schematic of active learning process.}
  \label{fig:Figure2}  
\end{figure}

In Bayesian neural networks, the constant weights are replaced with prior probability distributions (Figure \ref{fig:Figure2}). These priors are then updated through Bayes' rule to derive the posterior distributions after observing data. This modification addresses some of the limitations inherent in conventional neural networks, such as overfitting and the inability to quantify uncertainty in predictions. Generally, given the initial dataset \( \mathcal{D} \) comprised of inputs \( x \) and targets \( y \), an FBNN can be defined as a probabilistic model of the form:

\begin{equation}
  \begin{aligned}
    \text{Weights:} & \quad w \sim p(w) \\
    \text{Noise Variance:} & \quad \sigma^2 \sim \text{Half-Normal}(\sigma_0) \\
    \text{Predictions:} & \quad y \sim \mathcal{N}(g(x; w), \sigma^2)
  \end{aligned}
\end{equation}
where \( g \) is a neural network parameterized by its weights \( w \), and \( \sigma^2 \) is the observational noise variance. The idea of using prior distributions is to represent our initial beliefs about the model parameters before observing any data. However, in practice, defining meaningful priors for FBNN weights is highly non-trivial due to the large number of weights involved and the fact that these weights do not have an easily interpretable relationship with observable features of the data. Typically, generic priors such as normal distributions are employed, which do not necessarily convey specific prior knowledge but are chosen for their mathematical tractability and convenience. These priors serve as regularizers to prevent overfitting and allow for the incorporation of uncertainty into the model.

The HMC sampling technique \cite{hmc_intro} or its variants is then employed to approximate the posterior distribution of the weights. The predictive mean and uncertainty at new data points \( x^* \) are then computed as:
\begin{align}
\mu(x^*) &= \frac{1}{N} \sum_{i=1}^N g(x^*, w_{\text{post}}^{(i)}) \label{eq:mean_bnn} \\
U(x^*) &= \frac{1}{N} \sum_{i=1}^N (y_*^{(i)} - \mu(x^*))^2 \label{eq:uncertainty_bnn}
\end{align}
where \( y_*^{(i)} \sim \mathcal{N}(g(x^*, w_{\text{post}}^{(i)}), \sigma_{\text{post}}^{2(i)}) \) are samples drawn from the posterior predictive distribution and \textit{N} is the total number of HMC samples.

\subsection{Gaussian Process}

A Gaussian Process is considered the "gold standard" in active learning due to its flexibility and capacity for predictive uncertainty estimation \cite{rasmussenbook}. As such, it serves as the benchmark against which we compare FBNNs. The GP models a distribution over functions, making it highly adaptable to a wide range of applications.

The standard GP can be written as a probabilistic model of the form:

\begin{equation}
    \begin{aligned}
      \text{Kernel Hyperparameters:} & \quad \theta \sim p(\theta) \\
      \text{Noise Variance:} & \quad \sigma^2 \sim \text{Half-Normal}(\sigma_0) \\
      \text{Latent Function:} & \quad f(x) \sim \mathcal{GP}(0, k(x, x'; \theta)) \\
      \text{Observations:} & \quad y = f(x) + \epsilon, \quad \epsilon \sim \mathcal{N}(0, \sigma^2)
  \end{aligned}
\end{equation}

After training, the overall predictive mean at new points \( x^* \) is given by:
\begin{equation}
\mu(x^*) = \frac{1}{N} \sum_{i=1}^N \mu_{\text{post}}^{(i)} (x^*) \label{eq:gp_mean}
\end{equation}
where \( \mu_{\text{post}} \) is the predictive mean for a single HMC sample with kernel hyperparameters \( \theta \) and noise variance \( \sigma^2 \), defined using a textbook GP formula:
\begin{equation}
\mu_{\text{post}}(x^*) = k_*^T (K + \sigma^2 I)^{-1} y \label{eq:gp_formula}
\end{equation}
and where \( k_* \) is the vector of covariances between \( x^* \) and all training points \( x \), \( K \) is the covariance matrix of the training points, and \( I \) is the identity matrix.

Given that each individual GP produces its own uncertainty estimation, it is instructive to express the overall predictive uncertainty using the average of individual predictive variances and the variance across the predictive means of different HMC samples:
\begin{equation}
U(x^*) = \frac{1}{N} \sum_{i=1}^N \left( \Sigma_{\text{post}}^{(i)} (x^*) + (\mu_{\text{post}}^{(i)} (x^*) - \mu(x^*))^2 \right) \label{eq:gp_uncertainty}
\end{equation}
where
\begin{equation}
\Sigma_{\text{post}}(x^*) = k_{**} - k_*^T (K + \sigma^2 I)^{-1} k_* \label{eq:gp_variance}
\end{equation}
and where \( k_{**} \) represents the variance of the process at the new point \( x^* \). Note that when using the Maximum A Posteriori (MAP) estimate, which effectively corresponds to \(N=1 \), the predictive mean \( \mu(x^*) \) reduces to \( \mu_{\text{post}}(x^*) \) and the predictive uncertainty \( U(x^*) \) simplifies to \( \Sigma_{\text{post}}(x^*) \).

\subsection{Active Learning Strategy}

Active learning aims to efficiently utilize limited experimental or computational resources by iteratively selecting the most informative data points from a predefined parameter space. Each selected points corresponds to a parameter (1D case) or a combination of parameters (\textit{n}-D case) that the model predicts will yield the most significant insights about the system under study. The process begins with the selection of the overall parameter space, which is defined based on parameters that can be directly controlled and manipulated in experiments or simulations to affect behaviour of a system under study.

Starting with an initial dataset, usually represented by a small number of randomly or uniformly distributed measurements, the algorithm then selects data points from the parameter space based on the highest uncertainty in the prediction of surrogate model. The model is updated continuously as new data are acquired. This process is repeated until certain conditions are met, such as reaching a set number of measurements. This strategy ensures that resources are focused on exploring the most potentially valuable areas of the parameter space.

\begin{algorithm}[H]
\caption{Active Learning}
\begin{algorithmic}[1]
\State Define the overall parameter space \( \mathcal{X}_{\text{domain}} \)
\State Select the surrogate model type (FBNN or GP)
\State Initialize dataset \( \mathcal{D} = \{ (x_i, y_i) \}_{i=1}^N \)
\State Train the initial surrogate model on \( \mathcal{D} \)
\Repeat
    \State Compute uncertainty \( U(x^*) \) for each \( x^* \in \mathcal{X}_{\text{domain}} \)
    \State \( x_{\text{next}} \gets \arg\max_{x^* \in \mathcal{X}_{\text{domain}} } U(x^*) \)
    \State Conduct an experiment or simulation at \( x_{\text{next}} \) to obtain \( y_{\text{next}} \)
    \State Update \( \mathcal{D} \) by adding \( (x_{\text{next}}, y_{\text{next}}) \)
    \State Re-train the surrogate model on updated \( \mathcal{D} \)
\Until{a predefined stopping criterion is met (e.g. a number of measurements)}
\end{algorithmic}
\end{algorithm}

The goal of active learning is to optimize the efficiency and effectiveness of data acquisition in experimental or computational settings, conserving resources and potentially accelerating the discovery process.

\subsection{Choice of hyperparameters}
We standardized the hyperparameters across all tests for GP and FBNN models to ensure consistency in our comparisons. Specifically, for the Gaussian Process, we utilized a Mat\'ern 5/2 kernel with a weakly informative prior on the lengthscale, \( l \sim \text{Log-Normal}(0, 1) \). For the Bayesian Neural Network, unless explicitly specified otherwise, our architecture consisted of a three-layer fully-connected neural network with 32, 16, and 8 neurons in each respective layer, each activated by the hyperbolic tangent function, with normal priors over their weights, \( w \sim \text{Normal}(0, 1) \). In all models, unless explicitly specified otherwise, we adopted an observational noise prior of \( \sigma \sim \text{Half-Normal}(1) \), which is typical for scenarios where the variance of noise is expected to be moderate but should not dominate the signal.

\subsection{Model training}
To infer the posterior distributions over model parameters, we use the No-U-Turn Sampler (NUTS) \cite{hoffman2011nouturn}. It builds upon the HMC approach by automatically tuning the step size and dynamically setting the number of steps to take in each iteration. This automated tuning aspect of NUTS is particularly useful in autonomous science applications where manual tuning is infeasible or suboptimal. Here,  we utilized the version of NUTS from NumPyro probabilistic programming library \cite{phan2019numpyro} due to its compatibility with JAX’s \cite{jax2018github} automatic differentiation and just-in-time compiling capabilities for significantly faster computations. All computations were performed with double (64bit) precision on AWS EC2 t3.2xlarge instance.

\subsection{Test functions and datasets}

We have designed our test functions (Figure~\ref{fig:Figure3}a) to reflect common challenges in applying active learning to physical systems that are not always addressed by available benchmark functions.

\begin{itemize}
    \item \textbf{Discontinuous Transition 1:} This function models a system undergoing a discontinuous phase transition, starting with one power-law behavior and then abruptly switching to another at a specific point, representing a change from one phase to another.

    \item \textbf{Discontinuous Transition 2:} This function represents another potential example of phase transition dynamics. It transitions from a sinusoidal form to a sinusoidal with an added constant shift. This changes the baseline level of the function, leading to a discontinuity at that transition point.
    
    \item \textbf{Discontinuous Transition 3:}  This function models a system undergoing multiple discontinuous transitions, starting with a power-law increase, followed by moderated growth described by a polynomial of a lower degree and a negative offset, and concluding with exponential progression. It showcases how complex systems can exhibit multiple distinct states as external conditions change.
\end{itemize}

\begin{itemize}
    \item \textbf{Nonstationary 1:} This function is modified version of the higdon function \cite{gramacy2009adaptive} that transitions from a slowly varying oscillatory pattern to a linear trend, marking a clear shift in both the type and the nature of the function's output.

    \item \textbf{Nonstationary 2:} Designed to test the model's capability to handle abrupt changes in frequency and amplitude, this function defines a sinusoidal wave that switches to a high-frequency oscillation modulated by an exponential decay within a specified range.

    \item \textbf{Nonstationary 3:} Combining a predictable sinusoidal base with sporadic Gaussian spikes, this function mimics real-world phenomena where sudden, localized changes occur within a broader steady pattern, challenging the model's ability to predict under varying local conditions.
\end{itemize}

\begin{itemize}
    \item \textbf{Phases2D:} This function generates a two-dimensional domain mimicking a complex phase diagram. It tests the model's ability to discern and adapt to variations in physical and structural properties under different conditions which is essential in fields such as materials science.

    \item \textbf{Rays2D:} This function simulates the intensity of rays over a 2D domain, mimicking the study of magnetotransport phenomena \cite{qiu2020quantumhall} in condensed matter physics, which can produce ray-like patterns.
\end{itemize}

\begin{itemize}
    \item \textbf{Ising2D:} This function generates magnetization and heat capacity maps for 2D Ising model based on \cite{Kalinin_GP_Ising}, providing a test bed for active learning applications in physical simulations.
\end{itemize}

A small, zero-centered normally distributed noise is added to simulate observational noise. We start with 4 uniformly spaced points for 1D test functions and 10 randomly distributed points for three different random seeds for 2D test functions. For this study, we make a simplifying assumption that the system can be measured everywhere in the parameter space at any point in time, which may not be the case for non-reversible processes. We leave the exploration of models under constraints of irreversible changes as potential avenues for future research.

\section{Results and Discussion}\label{sec2}

\begin{figure}[h] 
  \centering  
  \includegraphics[width=1.0\textwidth]{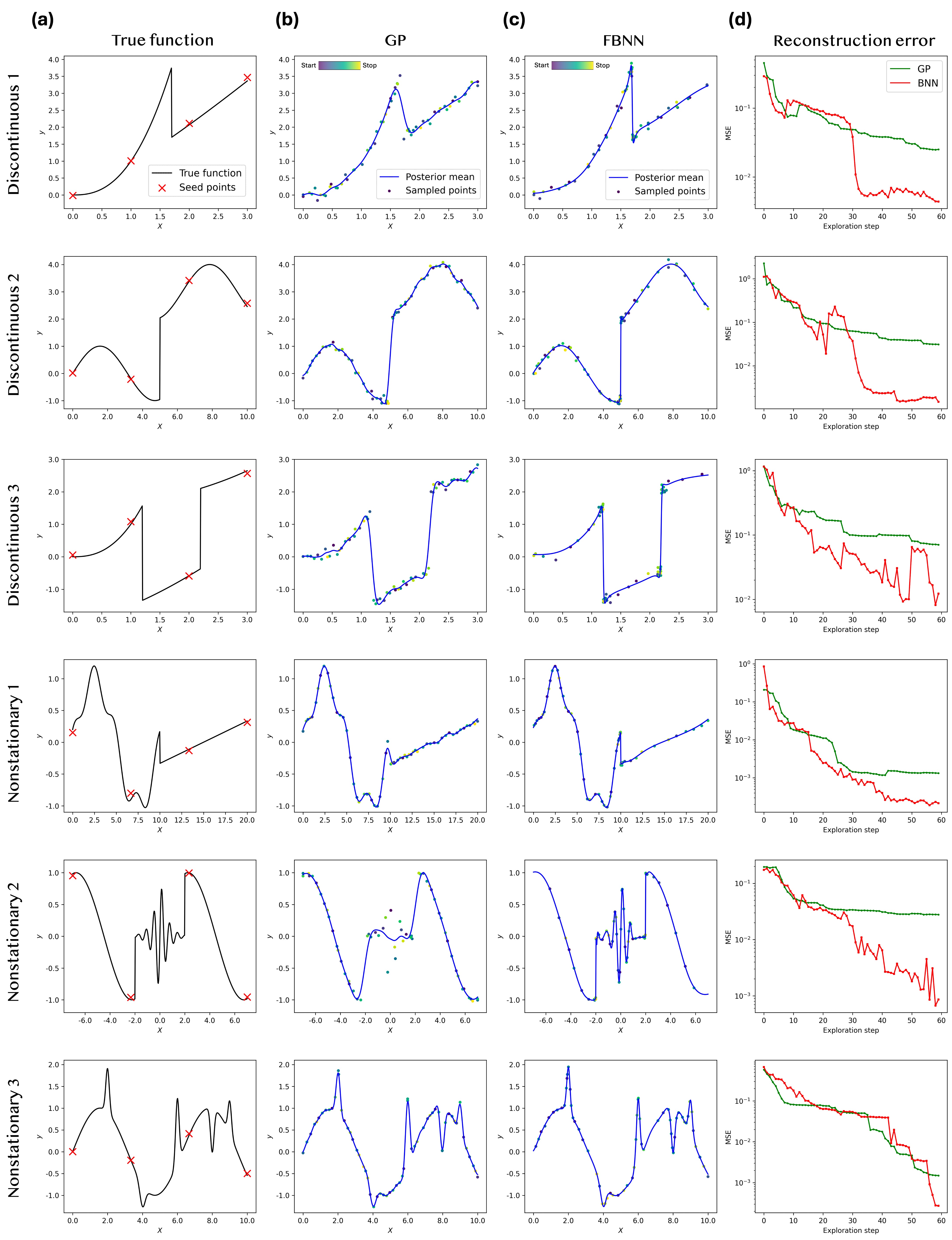}
  \caption{Active learning results for 1D test functions. (a) Ground truth function values with uniformly initialized starting ('seed') points. (b, c) Results of active learning with Gaussian process (GP) and Bayesian neural network (FBNN). (d) Mean squared error (MSE) as a function of active learning steps.}
  \label{fig:Figure3}
\end{figure}

The comparison of active learning results for 1D test functions is shown in Figure~\ref{fig:Figure3}. One can see that for the discontinuous functions, the GP tends to smooth over the transitions, lacking the ability to capture the abrupt shifts in the function. In contrast, FBNNs exhibit a higher flexibility, representing the sharp transitions with greater accuracy. For the non-stationary functions, the traditional GP, constrained by its global length scale assumption, struggles with specific behaviors such as straight lines or rapid high-frequency oscillations in the first two functions. Here, FBNN demonstrates greater adaptability. For the third nonstationary function, FBNN and GP provide comparable performance. Overall, while FBNNs may not consistently surpass GPs in the early stages of exploration, they tend to outperform GPs in later phases. We stress that all the FBNN results were obtained for the exact same network architecture, the same priors, and the same NUTS parameters. In other words, we did not tune neural network hyperparameters or the parameters of the training process for different test functions.

\begin{figure}[h]
  \centering
  \includegraphics[width=1.0\textwidth]{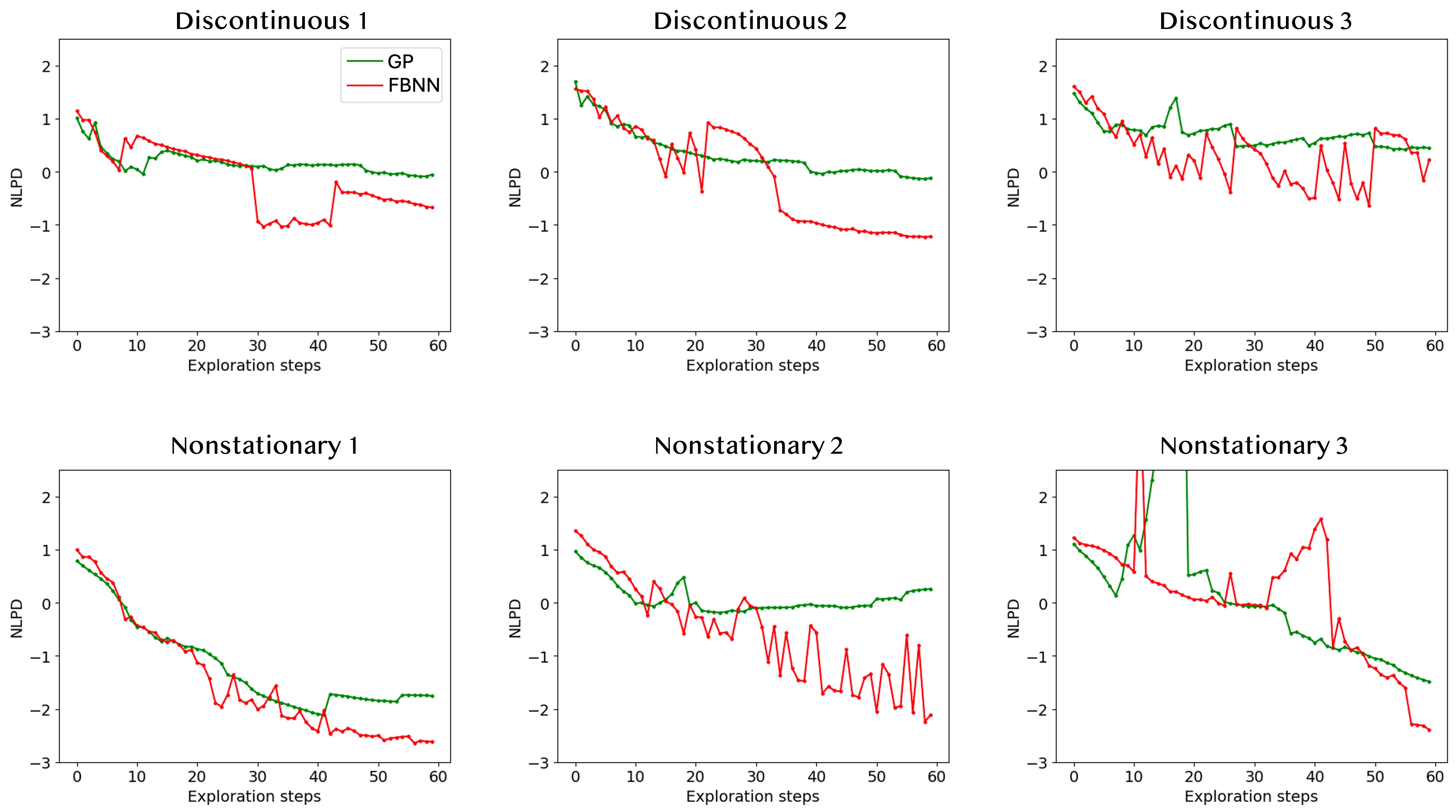}
  \caption{Perfomance evaluation of GP and FBNN in terms of negative log predictive density (NLPD) over active learning steps for the 1D test functions.}
  \label{fig:Figure4}
\end{figure}

Figure~\ref{fig:Figure4} examines the performance of FBNN and GP in terms of Negative Log Predictive Density (NLPD) over active learning steps. NLPD assesses the predictive accuracy of probabilistic models by measuring how well they predict unseen data, specifically focusing on their confidence in those predictions \cite{1202463_nlpd}. A model with a lower NLPD not only predicts more accurately but also indicates a more precise estimation of its uncertainty. FBNNs generally achieve lower NLPD values for both discontinuous and nonstationary function types, indicating better predictive accuracy and confidence. Notably, the FBNN curves tend to exhibit significant volatility, as their predictive confidence can fluctuate while adjusting to new information during the learning process. Despite these fluctuations, FBNNs generally stabilize at lower NLPD values, or even if they continue to fluctuate, their running averages tend to be smaller than those of GPs. 

\begin{figure}[h]
  \centering
  \includegraphics[width=1.0\textwidth]{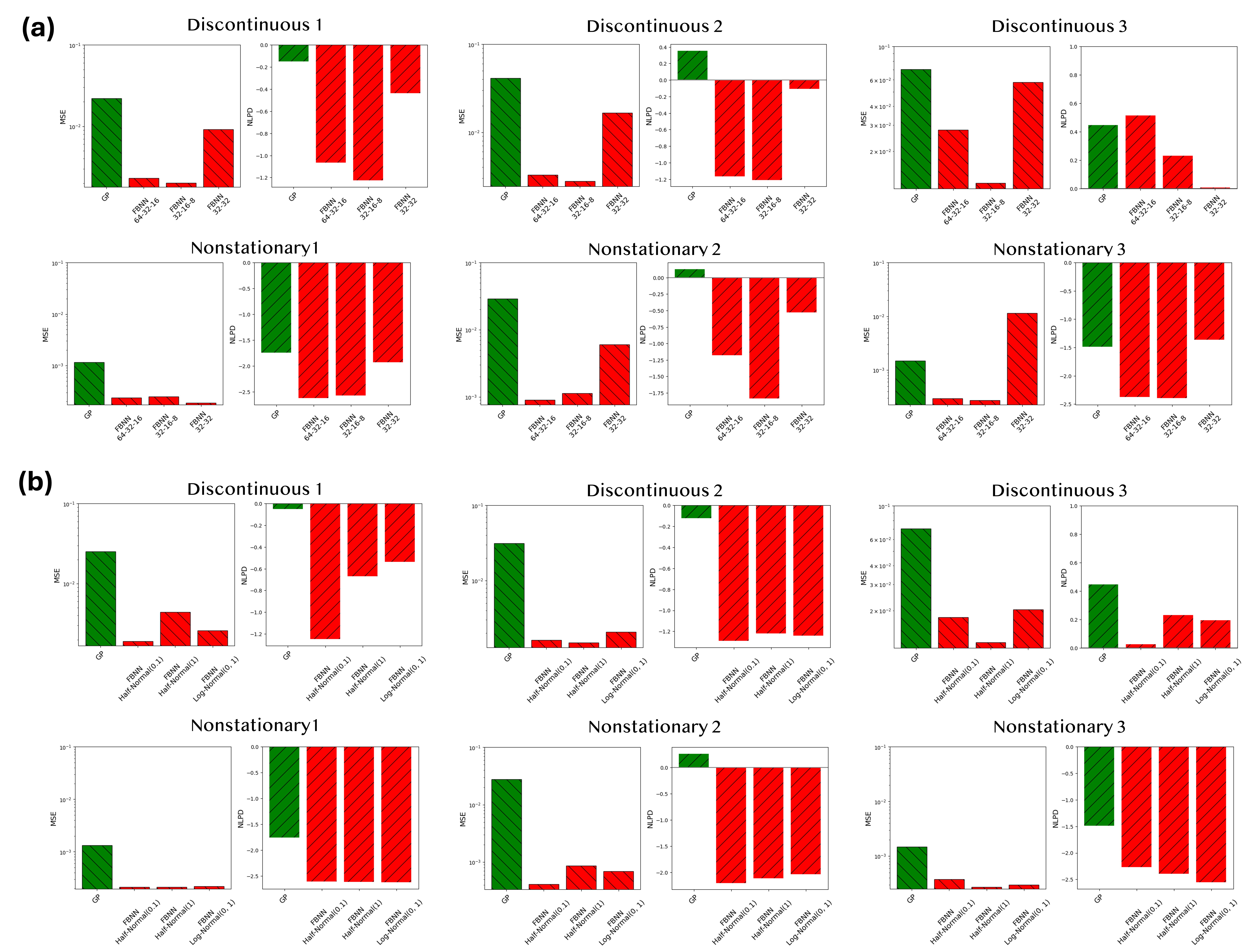}
  \caption{Performance evaluation in terms of MSE and NLPD at the final active learning step across (a) different FBNN architectures, (b) Different FBNN noise priors. The GP 'baseline' is shown by a green bar.}
  \label{fig:Figure5}
\end{figure}

The obvious question is how sensitive FBNN results are to the choice of neural network architecture. In Figure~\ref{fig:Figure5}a, we compared FBNN predictions (red bars) in terms of MSE and NLPD across three different architectures, as well as against the GP 'baseline' (green bar) at the last active learning step. Specifically, we assess the following FBNN architectures: a three-layer FBNN with 64, 32, and 16 neurons in successive layers (64-32-16), a three-layer FBNN with 32, 16, and 8 neurons (32-16-8), and a two-layer FBNN with 32 neurons in each layer (32-32). In terms of MSE, FBNNs outperform GPs in 5 out of 6 cases, regardless of the architecture. The only exception is the \textit{nonstationary-3}, for which a two-layer FBNN underperforms compared to the GP. For NLPD, FBNNs with different architectures either exceed or closely match the performance of GPs, except for the \textit{discontinuous-3}, where none of the models provide reliable performance (NLPD is positive in all cases).

In addition, for a Bayesian approach, the choice of observational noise prior can be important \cite{hafner2019bbbnoise}. In Figure~\ref{fig:Figure5}b, we analyze FBNN predictions using three different noise priors: \( \sigma \sim \text{Half-Normal}(0.1) \), \( \sigma \sim \text{Half-Normal}(1) \), and \( \sigma \sim \text{Log-Normal}(0, 1) \). Despite slight variations in MSE and NLPD values, FBNNs consistently outperform GP across all tested priors, underscoring the effectiveness of FBNNs for the discontinuous and nonstationary functions even under different noise assumptions. We didn't compare different weights priors since the normal distribution with zero mean and a standard deviation of one is already sufficiently generic. Overall, this capability of FBNNs to deliver competitive results even with potentially suboptimal configurations suggests that they offer a reliable alternative to traditional GPs.

\begin{figure}[h]
  \centering
  \includegraphics[width=1.0\textwidth]{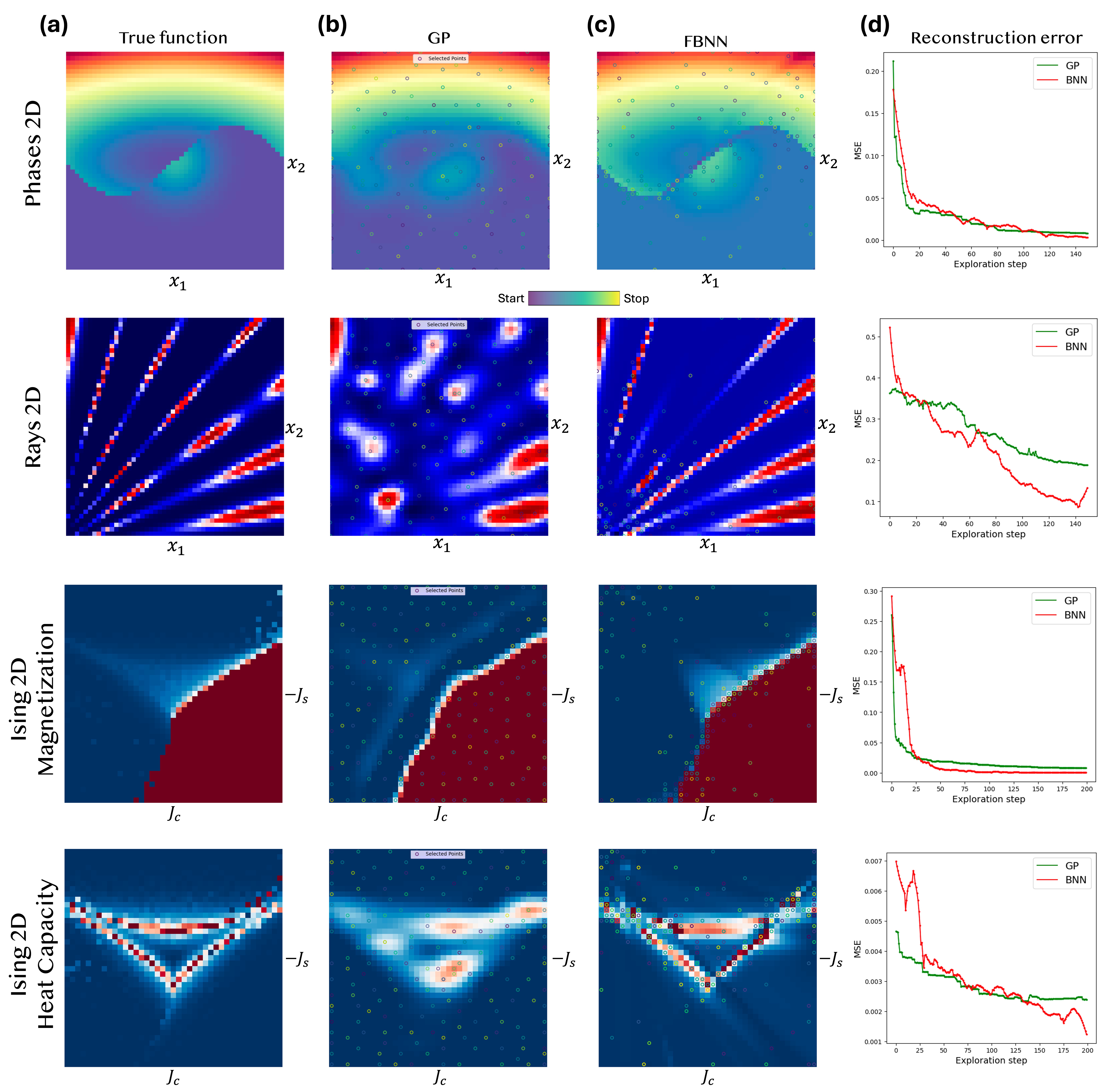}
  \caption{Active learning results for 2D test functions. (a) Ground truth function values. (b, c) Results of active learning with GP and FBNN (d) Mean squared error as a function of active learning steps. The reconstruction error was averaged over 3 runs with different initialization of starting points}
  \label{fig:Figure6}
\end{figure}

We next compare GP and FBNN based active learning performance for 2D parameter spaces (Figure~\ref{fig:Figure6}). For Phases2D, FBNN focuses on the phase boundary and avoids spending excessive time exploring regions of the phase space where values are nearly constant, demonstrating a more targeted and efficient exploration strategy. GP captures the overall landscape reasonably well but fails to precisely detect sharp phase boundaries, resulting in inefficient use of resources by focusing too much on areas with minimal variation. For Rays2D, the GP significantly underperforms, failing to adequately model the underlying data structure. By contrast, FBNN more accurately represents the pattern, clearly delineating the rays and capturing most of their intrinsic structure. 

Similarly, in the 2D Ising model, FBNN excels at delineating the phase boundaries for magnetization and effectively focuses on regions of the parameter space where the heat capacity is maximized. This clearly demonstrates its ability to identify key areas based purely on predictive uncertainty, without the need to introduce a tailored acquisition function as it was done for GPs in \cite{Kalinin_GP_Ising}. We emphasize that we used the same FBNN hyperparameters across all 2D test functions, and they were the same as those used for the 1D test functions in Figure~\ref{fig:Figure3}.

Given that FBNNs are at least several times more computationally expensive than GPs, and that GPs often capture the overall landscape quite well, one might consider whether it makes sense to perform active learning with GPs to obtain the overall landscape and then use a "one-shot" learning approach with FBNN to reconstruct finer details. We note, however, that a better delineation of transition and nonstationary regions by FBNNs is due to their focused active learning along those regions, where FBNN predictions tend to exhibit high uncertainty. This prompts the algorithm to allocate more resources to these areas. In contrast, the sampling strategies employed by GPs (or those based on random sampling) typically do not focus sufficiently on these regions. Consequently, a "one-shot" FBNN applied to data points collected by GP is unlikely to achieve the level of detail and accuracy that an FBNN-driven active learning process can provide. However, it would be interesting to explore if partially stochastic strategies \cite{pmlr-v206-sharma23a-partiallystochastic} for FBNN training could reduce computational cost and time without sacrificing predictive accuracy.

\section{Conclusions}
To summarize, we explored the performance of Fully Bayesian Neural Networks and Gaussian Processes in active learning contexts relevant to physical sciences problems. Our findings reveal that FBNNs outperform GPs in handling abrupt transitions and non-stationary behaviors, demonstrating robust adaptability and more accurate uncertainty quantification. Furthermore, FBNNs consistently demonstrated a competitive edge over GPs across multiple tested architectures and priors, highlighting their potential as a practical replacement for GPs, as they did not require extensive hyperparameter tuning.  Given the computational feasibility of FBNNs in lower-dimensional spaces and the demonstrated superior performance, they emerge as a compelling choice for active learning frameworks in autonomous science within "small data" regimes.  However, further exploration and application in experimental and modeling tasks are warranted, particularly to assess their performance in higher-dimensional spaces.

\section*{Acknowledgment}
This work was supported by the Laboratory Directed Research and Development Program at Pacific Northwest National Laboratory, a multiprogram national laboratory operated by Battelle for the U.S. Department of Energy. 

\section*{Code and Data Availability}
The source code for BNN, GP, and test functions is available at https://github.com/ziatdinovmax/NeuroBayes.

\bibliography{references}

\end{document}